\documentclass[letterpaper, 10 pt, conference]{ieeeconf}  

\IEEEoverridecommandlockouts                              

\usepackage[T1]{fontenc}
\usepackage{amsmath} 
\usepackage{amssymb}  
\usepackage{xcolor}
\usepackage{multirow} 
\usepackage{graphicx}
\usepackage{booktabs}
\usepackage{makecell}
\usepackage{array}
\usepackage{url}
\usepackage{tabularx} 

\usepackage[accsupp]{axessibility}  
\newcommand{\JC}[1]{\textcolor{black}{#1}}
\newcommand{\shan}[1]{\textcolor{black}{#1}}

\def\ours{\texttt{TrajPred}}

\title{\LARGE \bf
\JC{TrajPred}: Trajectory-Conditioned Joint Embedding Prediction for Surgical Instrument-Tissue Interaction Recognition \\in Vision-Language Models
}

\author{Jiajun Cheng, Xiaofan Yu, Subarna Tripathi, Sainan Liu, Shan Lin}

\begin{document}

\maketitle
\thispagestyle{empty}
\pagestyle{empty}

\begin{abstract}
\shan{Recognizing instruments' interactions with tissues} is essential for building context-aware AI assistants in robotic surgery. Vision-language models (VLMs) have opened a new avenue for surgical perception \shan{and achieved better generalization on a wide range of tasks compared to conventional task-specific deep learning approaches. However, their performance on instrument–tissue interaction recognition remains limited, largely due to two challenges:} (1) Many models do not effectively leverage temporal information, (2) alignment between vision and text often misses fine-grained action details. To address these issues, we propose \ours{}, a framework that encodes instrument trajectories to incorporate temporal motion cues; and conditioned on these trajectories, introducing a predictor module to generate visual semantic embeddings that better capture fine-grained action details.
We further incorporate prompt tuning and a verb-rephrasing technique to enable smooth adaptation \shan{to the Instrument-Tissue interaction recognition task}. Extensive experiments on \shan{the public laparoscopic benchmark, CholecT50,} show that our method improves both Average Precision and Top-K accuracy. 

\shan{We also investigate whether visual embeddings of instrument-tissue interaction regions align better with the corresponding text by visualizing the cosine similarity between visual and textual embeddings. The visualization results indicate that the proposed method improves alignment between relevant visual and textual representations. Code is available at \url{https://github.com/jiajun344/TrajPred}.}
\end{abstract}

\section{Introduction} \label{sec:intro}


\shan{Robotic surgery has been widely adopted due to improved ergonomics, clinical outcomes, and reduced complication rates~\cite{dupont2021decade}. However, in hospitals, current robotic systems remain
fully surgeon-operated, and surgical quality still depends heavily on surgeon experience,
performance variability, and availability. These limitations have motivated increasing interest in robotic automation to achieve more consistent and reproducible
performance~\cite{yip2025robot, schmidgall2025will}. 
Recognizing Instrument-Tissue interactions is a critical component for enabling effective collaboration between robots and surgeons, allowing robots to understand and respond appropriately to surgeons’ actions. In the future, this capability can also guide robots in learning surgical skills from expert demonstrations.}

\begin{figure}[t]
\centering
\includegraphics[width=1.0\columnwidth]{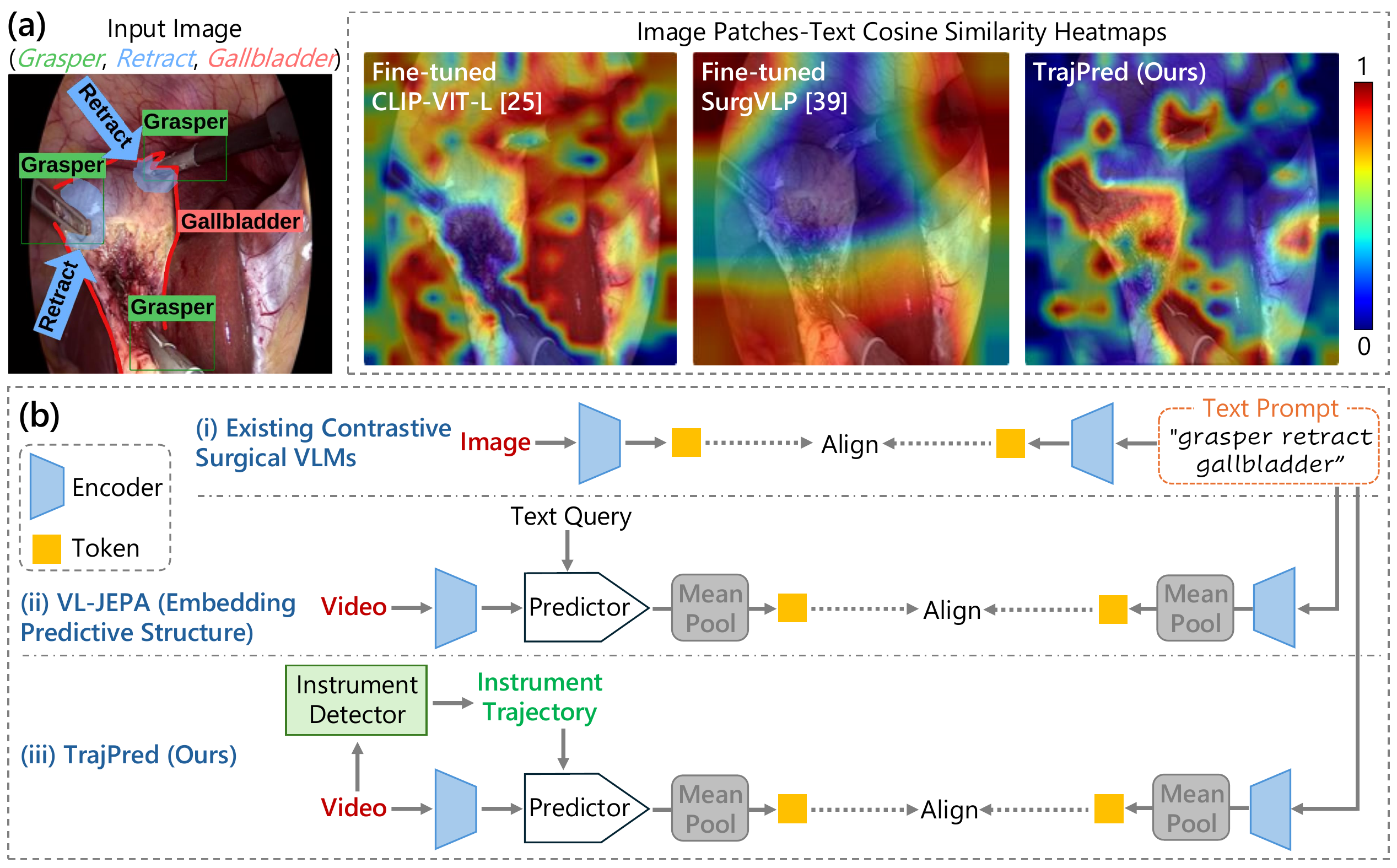 }
\vspace{-6mm}
\caption{(a) Cosine similarity heatmaps between embeddings of image patches and the text (e.g., “grasper retract gallbladder”). Two representative contrastive learning-based VLMs (CLIP-ViT-L \cite{radford2021learning} and SurgVLP \cite{yuan2023surgvlp}) exhibit high similarity between background patches and the text, whereas \ours{} (ours) produces embeddings that concentrate on the Instrument-Tissue interaction region. 
(b) Comparison of the architectures of (i) popular contrastive learning-based surgical VLMs, (ii) VL-JEPA\cite{chen2025vl}, and (iii) \ours{}.}\label{fig_intro}
\vspace{-8mm}
\end{figure}

\shan{Instrument-Tissue interaction} recognition are typically formulated as surgical action triplets structured as (\textit{instrument}, \textit{verb}, \textit{target})\cite{nwoye2022data}.
Such triplet representations are also commonly used in general-domain action recognition tasks \cite{damen2018scaling}. Traditional surgical triplet recognition methods are typically built on classifier-based architectures that predict the instrument, verb, and target separately using independent classification heads \cite{nwoye2022rendezvous,liu2024surgical,gui2024tail}. Although these approaches have improved performance on commonly used surgical action triplet datasets \cite{nwoye2022data}, classifier-based architectures rely on clearly labeled samples, learning decision boundaries tailored to predefined training categories. However, obtaining such detailed annotations is expensive, and models trained in this way often struggle to generalize to diverse real-world surgical settings where procedures may appear under different environments or conditions. 


\JC{Recent progress in contrastive learning vision–language models (VLMs) offers a promising approach to addressing these generalization limitations. Unlike classifier-based architectures that learn to directly map images to predefined labels, VLMs learn representations by aligning visual features with textual descriptions in a shared embedding space~\cite{li2022blip_duplicate,radford2021learning}.
Pretrained on large-scale and diverse data, VLMs effectively bridge visual and natural language embeddings, this allows training to leverage audio captions from large publicly available surgical video databases online, reducing the labeling burden while learning semantic representations better generalize to the same surgical procedures under different environments and visual conditions.
Building on this paradigm, several works~\cite{yuan2023surgvlp,yuan2024hecvl,yuan2024peskavlp,zeng2025surgvlm} have leveraged large-scale surgical video datasets to develop VLMs
tailored to the surgical domain, demonstrating the potential of VLMs across various surgical perception tasks.} However, existing studies on surgical VLMs indicate that their performance on surgical action triplet recognition remains limited~\cite{rau2025systematic,zeng2025surgvlm}. 
Two key limitations hinder the performance of VLMs for Instrument-Tissue interaction recognition




\begin{figure*}[t]
\centering
\includegraphics[width=1.6\columnwidth]{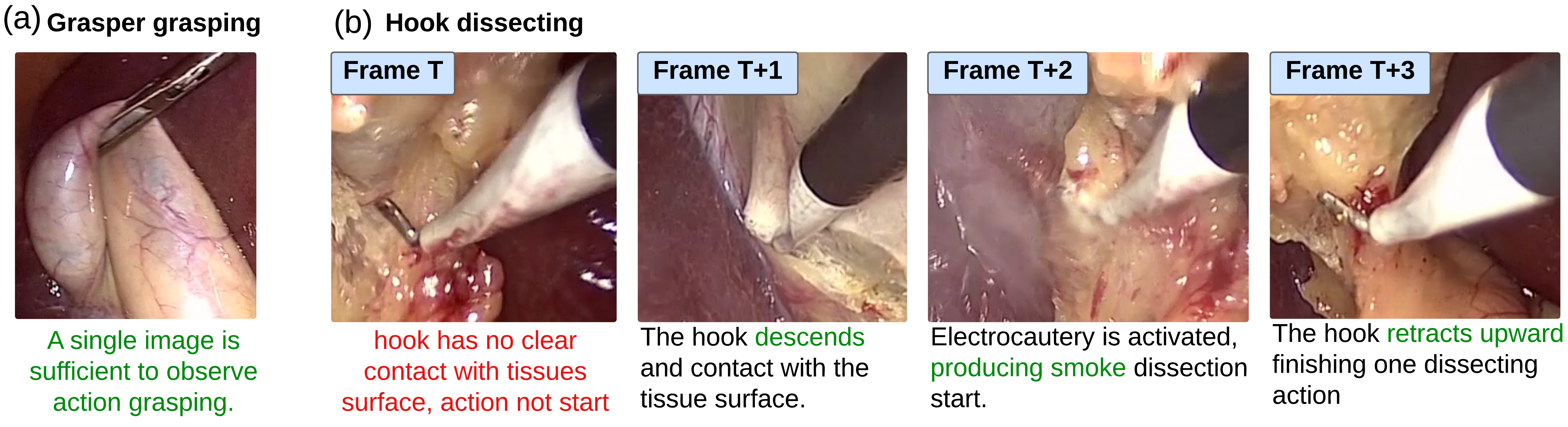}
\vspace{-4mm}
\caption{Comparison between frames of the actions (a)``Grasper grasp'' and (b)``Hook dissect''.}
\label{fig_verb}
\vspace{-6mm}
\end{figure*}

(1) \textbf{Limited Use of Detailed Temporal Information Across Frames.} 
Many existing surgical contrastive learning VLM approaches~\cite{sharma2025fine,yuan2023surgvlp,yuan2024hecvl,yuan2024peskavlp,zeng2025surgvlm} rely on single-frame inputs paired with text captions during training. Similar to traditional methods, many Instrument-Tissue interaction recognition models also operate on single frames~\cite{nwoye2022rendezvous,gui2024tail}.
Inferring surgical actions from a single frame is inherently ambiguous. While certain actions, such as grasping (Fig.~\ref{fig_verb}(a)), may be inferred from static visual cues, many others require information across multiple frames to observe motion patterns (Fig.~\ref{fig_verb}(b)). 
Although single-image models may achieve reasonable performance because certain instruments are strongly associated with specific actions, this reliance can limit the model’s ability to understand how instruments actually perform different actions. As a result, such approaches are less reliable for intraoperative applications, where accurate, robust, and explainable recognition of ongoing surgical actions is critical.


(2) \textbf{Suppression of Details in Contrastive Learning.} 
\JC{As current visual encoders commonly split images into multiple image patches, most surgical VLMs are trained using contrastive learning that aligns visual features aggregated from the entire image with the corresponding text (Fig. ~\ref{fig_intro}(b)(i)).} 
While effective for global semantics, this strategy often suppresses fine-grained spatial details~\cite{choi2025goal,yao2021filip}, \JC{but in a surgical setting, pooled visual features may contain substantial background information, with only a small portion relevant to the text.} 
Fig.~\ref{fig_intro}(a) shows an example comparing image patch-text embedding cosine similarity heatmaps between two representative contrastive learning-based VLMs (surgical-pretrained SurgVLP~\cite{yuan2023surgvlp} and the general pretrained CLIP~\cite{radford2021learning}) and our \ours{}, where existing models highlight background whereas our model focuses on the Instrument-Tissue interaction region. 
Recent surgical VLM benchmark studies \cite{cheng2025surgxbench} have also highlighted the loss of fine-grained details and raised concerns regarding the reliability of current VLMs for surgical decision support. Although recent work such as \cite{choi2025goal,yao2021filip} proposes object-level alignment by aligning image patches fall within bounding boxes of target objects to corresponding text tokens, this approach is not directly applicable to our task, as Instrument-Tissue interaction involves action recognition and it is difficult or even impractical to define a bounding box for the action.


To mitigate these two limitations, we propose \ours{}, a trajectory-conditioned joint embedding prediction framework for vision–language. 
To better leverage temporal information, we use video clips as input and explicitly encode instrument trajectories to support action triplet recognition (Fig.~\ref{fig_intro}(b).(iii)). 
To mitigate detail suppression in contrastive learning, inspired by VL-JEPA~\cite{chen2025vl} (Fig.~\ref{fig_intro}(b).(ii); a VLM achieves state-of-the-art video understanding), we adapt its trainable predictor module to predict semantic embeddings conditioned on trajectory tokens, enabling better capture of fine-grained action details (Fig.~\ref{fig_intro}(b)(iii)).  Finally, fine-tuning VLMs often degrades their generalization ability \cite{chen2025prompt,zhou2022conditional}. To mitigate this issue, we reformulate surgical verbs into simplified but detailed english sentences and apply CoOp-style prompt tuning~\cite{zhou2022learning} to retain semantic priors in the pretrained text encoder for smooth Instrument-Tissue interaction recognition task adaptation. In summary, our contributions are as follows:
\begin{itemize}

\item \JC{We reformulate Instrument-Tissue interaction recognition as an embedding prediction problem, replacing contrastive learning alignment with a semantic embedding predictive framework for to capture fine-grained action details.}

\item We extract instrument trajectory tokens to condition the embedding prediction, capturing their temporal motion cues across frames for better action recognition.

\item \JC{We reformulate surgical verbs into descriptive prompts and apply prompt tuning to improve generalization capability.}
\item We conduct extensive experiments to demonstrate that \ours{} improves action triplet recognition performance and generalization while enhancing alignment between visual and text embeddings.

\end{itemize}

\section{Related Work}

\paragraph{\textbf{\JC{Instrument-Tissue Interaction recognition}}}


In the surgical domain, the Instrument-Tissue interaction recognition task aims to predict a structured representation of surgical activity in the form of an $(\text{instrument}, \text{verb}, \text{target})$ triplet~\cite{nwoye2022data}, e.g., (scissors, cut, tissue), indicating which instrument performs which action on which target. 
Traditional approaches typically adopt classifier-based frameworks that use separate classification heads for instrument, verb, and target~\cite{nwoye2022rendezvous,sharma2023rendezvous,liu2024surgical,gui2024tail}. However, such methods rely heavily on expensive well-labeled datasets and often struggle to generalize to the same surgical procedures performed under different environments or conditions.

\paragraph{\textbf{Surgical VLMs}}
VLMs have recently advanced surgical scene understanding. 
Prior surgical VLMs, including SurgVLP~\cite{yuan2023surgvlp}, HecVLP~\cite{yuan2024hecvl}, and PeskaVLP~\cite{yuan2024peskavlp}, leverage large-scale surgical video corpora for contrastive learning pretraining and domain adaptation, incorporating hierarchical supervision and LLM-augmented knowledge. 
Beyond image–caption contrastive learning, recent works~\cite{perez2025surglavi,honarmand2024vidlpro,wang2025surgvidlm,li2025surgpub} pretrain surgical video–language models by constructing large-scale surgical lecture video datasets collected online, adopting either contrastive learning objectives or autoregressive LLM-based training paradigms. Overall, current surgical VLM research primarily emphasizes dataset curation, focusing on the scale and quality of video–caption pairs. However, existing methods often overlook whether visual and textual representation are truly aligned in detailed-level, as shown in Fig.~\ref{fig_intro}(a), which is crucial for AI trustworthiness, particularly in the medical domain \cite{cheng2025surgxbench}. In this paper, we show that our \ours{} framework avoids overlooking detailed action recognition and achieves better visual feature–text embedding alignment.

\paragraph{\textbf{Temporal information in Surgery}}


Temporal information has been widely used in surgical data analysis tasks, such as semantic segmentation \cite{fang2026spatio,lin2021multi}, 
phase recognition \cite{gao2021trans,liu2025lovit}, and surgical action triplet recognition \cite{sharma2023rendezvous}. Most of these methods are still trained with a small dataset and focus on specific tasks without good generalization capability. In the VLM domain, most approaches on surgical VLM training with temporal information mainly rely on large-scale publicly available surgical video clips, pairing audio-transcribed text with video clips for VLM training \cite{perez2025surglavi,honarmand2024vidlpro,wang2025surgvidlm,li2025surgpub} without explicitly care about fine-grained details in surgical actions. In \ours{}, we address these issues by explicitly encoding detailed instrument trajectories to better capture fine-grained visual details.

\paragraph{\textbf{Fine-grained details alignment in VLMs}}
Recent works extend contrastive VLM alignment from global image features to finer spatial regions. 
Some methods use object detectors to align visual features from object regions with corresponding text tokens~\cite{chen2024contrastive,zeng2021multi,kamath2021mdetr}, while others apply soft region grouping to form coherent feature clusters and perform region–text contrastive learning without explicit localization~\cite{sharma2025fine,zohra2025beta,xu2022groupvit}. 
More recent approaches explore automatic object alignment, e.g., using SAM to extract object regions followed by pretrained contrastive models for object–sentence matching~\cite{choi2025goal}. 
Although these methods improve object-level alignment and model performance, surgical action understanding involves temporal relationships between objects ~\cite{liang2023relation,wang2023paxion} rather than merely forcing features within regions of objects in the video to align with their corresponding object texts
To address this, \ours{} leverages all visual features and employs a VL-JEPA-style predictor conditioned on trajectory information to learn action-aware semantic embeddings instead of relying on object-level alignment.



\section{Proposed Methods} \label{sec:method}

\JC{The proposed \ours{} framework reformulates conventional contrastive VLM training into an embedding prediction paradigm for the Instrument-Tissue interaction recognition task. Specifically, we replace standard contrastive alignment with the VL-JEPA-inspired predictive embedding structure~\cite{chen2025vl}, and further extend it with a trajectory encoding module to explicitly model instrument motion. Surgical action recognition task is commonly formulated as instrument–verb–target triplet recognition \cite{nwoye2022data}, e.g., $(\textit{hook}, \textit{dissect}, \textit{gallbladder})$, the triplet explicitly captures tool–tissue interactions. Similar structured representations are also widely used in general-domain egocentric action datasets \cite{damen2018scaling}.
We first describe the overall predictor architecture of VL-JEPA, which our method builds upon, in Section~\ref{sec:overall}, followed by how our \ours{} method integrates trajectory tokens in Section~\ref{sec:traj_token} and text encoder design and verb rephrasing method in Section~\ref{sec:text_encoder} and ~\ref{sec:verb_phrase}. Finally, we present the optimization and training objective in Section~\ref{sec:loss}.}

\subsection{VL-JEPA Structure}\label{sec:overall}

VL-JEPA~\cite{chen2025vl} consists of three components:
(i) a frozen visual encoder,
(ii) a predictor module, and
(iii) a text encoder.

\paragraph{Visual Encoder}
Given a video clip $\mathbf{V} = \{ I_t \}_{t=1}^{T}$, 
The visual encoder partitions it into spatio-temporal tubelet tokens and produces 
$\mathbf{Z}_v \in \mathbb{R}^{N_v \times D_v}$. 
The visual encoder remains frozen during training.

\paragraph{Text Encoder}
The text encoder maps a natural language query into a semantic embedding 
$\mathbf{z}_t \in \mathbb{R}^{D_t}$, 
which serves as the alignment target in the joint embedding space. The text encoder is trainable during training.

\paragraph{Predictor}
The predictor takes all visual tokens representation $\mathbf{Z}_v$ and query tokens to predict final target embedding, unlike contrastive learning methods that rely on a single \texttt{[CLS]} token that summarize entire clip as the final visual embedding. The final embedding $h \in \mathbb{R}^{D_t}$ is obtained via mean pooling over the predictor outputs.

\begin{figure}[t]
\centering
\includegraphics[width=1.0\columnwidth]{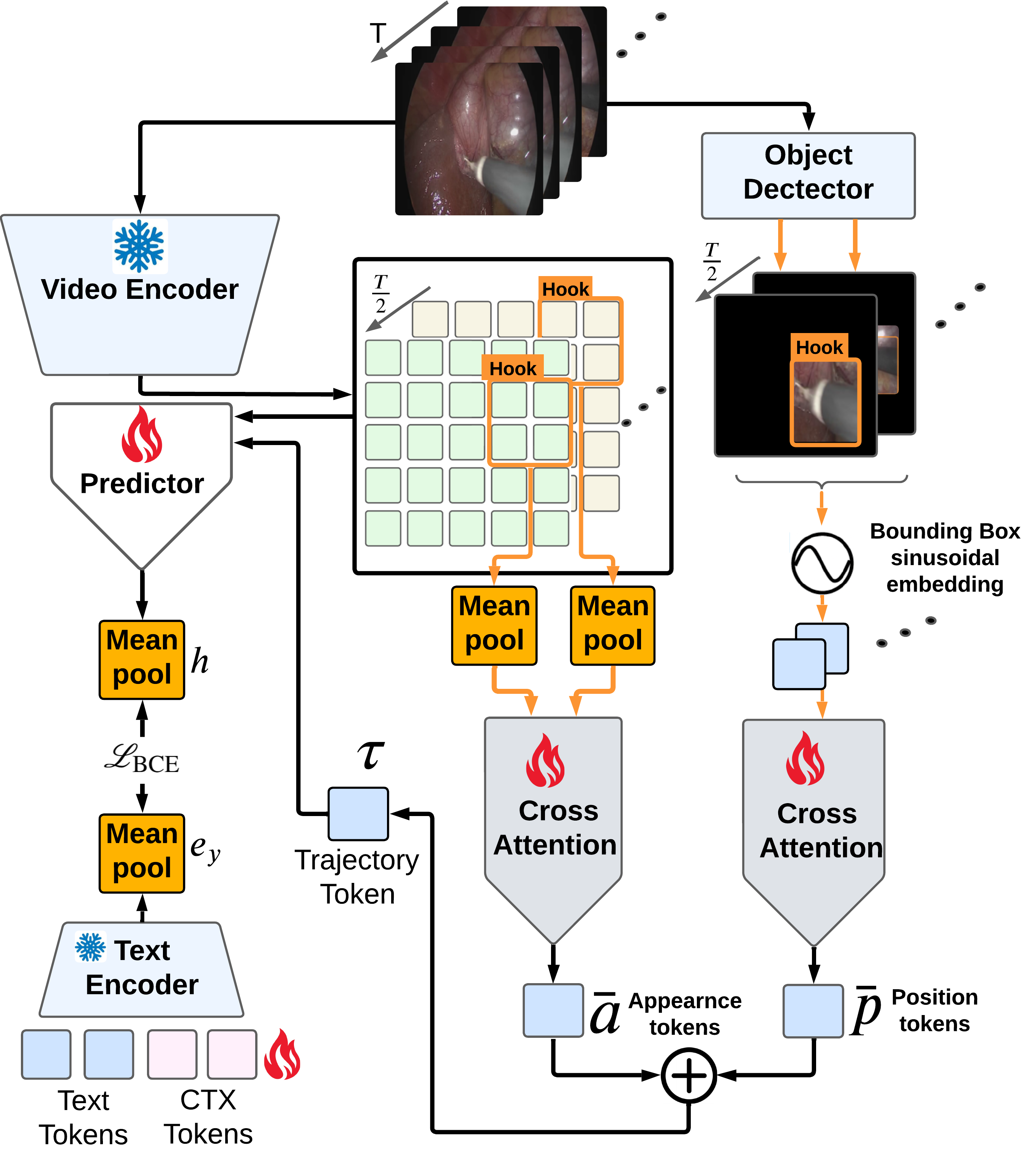}
\vspace{-8mm}
\caption{Overview of \ours{} pipeline. Video clips are processed by both the video encoder and an object detector. Visual tokens within detector-predicted bounding boxes are mean-pooled and combined with bounding box coordinate embeddings through element-wise addition, forming trajectory tokens that are subsequently encoded via cross-attention. The resulting trajectory tokens are jointly fed with video encoder tokens into the predictor.}
\label{fig_pipline}
\vspace{-6mm}
\end{figure}

\subsection{\ours{} Framework}\label{sec:traj_token}

\JC{ 
\JC{While VL-JEPA demonstrates strong performance in video understanding tasks through embedding prediction and predictor module leverage all visual encoder outputs tokens~\cite{chen2025vl}, surgical videos often contain significant spurious motion (e.g., camera movement), which may still lead the predictor to attend to spurious cues. We therefore introduce a trajectory token encoding pathway to explicitly guide learning toward informative instrument dynamics.}
Inspired by \cite{zheng2025one}, we construct a single trajectory token for each detected instrument $k$ in the clip. 
The method consists of two streams: (1) an appearance stream that captures the instrument’s visual features across frames, and (2) a position stream that captures the instrument’s trajectory information over time,} as illustrated in Fig.~\ref{fig_pipline}. We first employ a Fast R-CNN–based detector~\cite{jin2018tool} trained on surgical instrument data to extract bounding boxes for each instrument in every frame\JC{, the bounding boxes serve as positional information for constructing trajectory tokens, as instrument movements—toward, away, or in different directions—are reflected in the changes of their bounding box coordinates.}
Notably, our framework is detector-agnostic, any object detection or instance segmentation model capable of localizing surgical instruments can be used to obtain the bounding box coordinates. 
Let $\{b_{t}^{(k)}\}_{t=1}^{T_k}$ denote the bounding box coordinates of instrument $k$ over $T_k$ frames.

\paragraph{Appearance stream}
For each frame $t$, we mean-pool tubelet tokens $\mathbf{Z}_v$ within the bounding box $b_t^{(k)}$ to obtain a per-frame appearance feature $a_t^{(k)} \in \mathbb{R}^{D_v}$. The temporal sequence $\{a_t^{(k)}\}_{t=1}^{T_k}$ is aggregated by a lightweight cross-attention pooling module (two layers, eight heads) with a single learnable query token that attends to the sequence as keys and values, producing
$
\bar{a}^{(k)} = \text{CrossAttnPool}(\{a_t^{(k)}\}_{t=1}^{T_k}),
$
where $\bar{a}^{(k)} \in \mathbb{R}^{D_v}$ summarizes the temporal appearance of instrument $k$. The position stream uses the same architecture with a separate learnable query.

\paragraph{Position stream}
\JC{The bounding box coordinates $b_t^{(k)} \in \mathbb{R}^{4}$ are mapped into $D_v$-dimensional sinusoidal positional embeddings $p_t^{(k)} \in \mathbb{R}^{D_v}$ using fixed sine and cosine functions at multiple frequencies, encoding the spatial location of each instrument at each timestep. The temporal sequence $\{p_t^{(k)}\}_{t=1}^{T_k}$ is then compressed by a separate same cross-attention pooling module as appearances stream with its own learnable query token into a single position token $\bar{p}^{(k)} \in \mathbb{R}^{D_v}$.}

\paragraph{Fusion}
The final trajectory token for instrument $k$ is defined as
$
\tau^{(k)} = \bar{a}^{(k)} + \bar{p}^{(k)} \in \mathbb{R}^{D_v}.
$
Element-wise summation is used to ensure each instrument trajectory is represented by a single token to ensure the token count is within a reasonable range. The trajectory tokens $\{\tau^{(k)}\}_{k=1}^{K}$ are concatenated with the video tubelet tokens $\mathbf{Z}_v$ to form
$
\mathbf{Z}_{\text{aug}} = \big[ \mathbf{Z}_v \, ; \, \tau^{(1)}, \dots, \tau^{(K)} \big],
$
where $K$ denotes the number of valid instrument trajectories. The augmented sequence $\mathbf{Z}_{\text{aug}}$ is then fed into the predictor.

\begin{table}[t]
\centering
\caption{Verb rephrasing used for prompt construction.}
\label{tab:verb_rephrase}
\vspace{-2mm}
\begin{tabularx}{\linewidth}{>{\centering\arraybackslash}m{4cm} X}
\toprule
\textbf{Verb} & \textbf{Rephrased Phrase} \\
\midrule
grasp      & holding and gripping \\
retract    & pulling aside \\
dissect    & separating by cutting \\
coagulate  & stopping bleeding by heating \\
clip       & clipping closed \\
cut        & cutting through \\
aspirate   & sucking fluid from \\
irrigate   & washing with liquid \\
pack       & pressing material onto \\
null\_verb & not acting \\
\bottomrule
\end{tabularx}
\vspace{-7mm}
\end{table}

\subsubsection{Text Encoder Tuning}\label{sec:text_encoder}
\JC{We follow the same text encoder design as VL-JEPA~\cite{chen2025vl}, obtaining full surgical action triplet embeddings using a pretrained Embedding Gemma encoder~\cite{vera2025embeddinggemma}. Instead of fine-tuning the entire text encoder, we adopt CoOp-style prompt tuning~\cite{zhou2022learning} for better generalization capability. This approach optimizes a small set of learnable context tokens while keeping the backbone model parameters frozen, thereby retaining more of the pretrained knowledge.} We prepend $M$ learnable context tokens $\{c_m\}_{m=1}^{M}$ to the tokenized text sequence $\{y_j\}_{j=1}^{L}$, forming the input $[c_1, \dots, c_M, y_1, \dots, y_L]$. Only the context tokens are trainable, while the Gemma encoder weights remain fixed. 
The encoder outputs are mean pooled final text embedding $e_y \in \mathbb{R}^{D_t}$.

\subsubsection{Verb Phrase}\label{sec:verb_phrase}
\JC{
To mitigate the domain gap between surgical-specific verbs and the general-language vocabulary learned during large-scale pretraining, 
we replace the original verb labels with more descriptive natural language phrases in both training and evaluation. 
The complete verb-to-phrase mapping is provided in Table~\ref{tab:verb_rephrase}, and all \ours{} results are reported using this formulation.
}

\subsection{Training Optimization}\label{sec:loss}
Let $h_i \in \mathbb{R}^{D_t}$ and $e_c \in \mathbb{R}^{D_t}$ denote the predictor output and text embedding for sample $i$ and class $c$, respectively. We compute cosine similarity logits $s_{i,c}$ and optimize a multi-label binary cross-entropy loss~\cite{zhou2022learning}:
\[
\mathcal{L}_{\text{BCE}}
=
\frac{1}{NC}
\sum_{i,c}
\Big[
- y_{i,c} \log \sigma(s_{i,c})
- (1 - y_{i,c}) \log \big(1 - \sigma(s_{i,c})\big)
\Big],
\]
where $y_{i,c} \in \{0,1\}$ is the ground-truth label and $\sigma(\cdot)$ is the sigmoid function.


\section{Experiments}

\subsection{Dataset and Metrics}

In this paper, we conduct experiments on the CholecT50 dataset\cite{nwoye2022data}, a surgical action triplet recognition benchmark consisting of 50 laparoscopic cholecystectomy videos sampled from CholecT80 video dataset \cite{twinanda2016endonet} and annotated in 1FPS with 100 action triplet classes in the form 〈instrument, verb, target〉. The dataset contains 100.9K video frames and over 151K triplet instances, constructed from 6 instrument classes, 10 verb classes, and 15 target classes. For reproducibility, we follow the official Rendezvous (RDV) split for training and testing.
Performance is evaluated using two metrics:
\paragraph{Average precision}
We report the Average Precision (AP) metrics following the official evaluation protocol introduced in CholecT50 dataset \cite{nwoye2022data}, including triplet-level AP$_{IVT}$ as well as component-level AP for instrument (AP$_I$), verb(AP$_V$), and target(AP$_V$).

\paragraph{Top-K accuracy} 
We report Top-$K$ accuracy to evaluate the ranking quality of triplet predictions. 
For each sample $i$, the model computes cosine similarity scores with all $C$ triplet prompts, producing a score vector $p_i \in \mathbb{R}^{C}$. 
Let $\text{TopK}(p_i)$ denote the top-$K$ highest-scoring triplets and $\mathrm{GT}_i$ the set of ground-truth triplets. 
The Top-$K$ accuracy over $N$ test samples is defined as
$
\text{Top@K}
=
\frac{1}{N}
\sum_{i=1}^{N}
\frac{
\left|
\text{TopK}(p_i) \cap \mathrm{GT}_i
\right|
}{
|\mathrm{GT}_i|
}.
$
This measures the proportion of ground-truth triplets retrieved within the top-$K$ predictions. 
We additionally report $\text{TOP}@(K = |\mathrm{GT}|)$, where $K_i = |\mathrm{GT}_i|$ is set to the number of ground-truth triplets for each sample and the final score is averaged over all samples.

\subsection{Evaluation Strategy}\label{eval_stra}

\paragraph{Baseline}
We compare \ours{} with seven baselines trained under the same CholecT50 RDV split, including CLIP~\cite{radford2021learning}, CLIP-GOAL~\cite{choi2025goal}, and surgical pretrained VLMs such as SurgVLP~\cite{yuan2023surgvlp}, HecVL~\cite{yuan2024hecvl}, and PeskaVLP~\cite{yuan2024peskavlp}. We also implement and follow the same setup as VL-JEPA~\cite{chen2025vl} with both image-based and video-based variants using the text query “What is the action shown?”

\paragraph{Experiment setting}We evaluate Performance under two settings.
(1) We follow the CholecT50 RDV split~\cite{nwoye2022data} for training and evaluation.
Inspired by~\cite{sharma2025fine}, we adopt an unseen-verb setting to evaluate generalization in surgical triplet prediction. 
Since the original split of~\cite{sharma2025fine} is not publicly available, we construct our own unseen-verb subset under the RDV split by removing all training frames containing the verbs \emph{irrigate}, \emph{retract}, \emph{pack}, and \emph{aspirate}, and evaluating on triplets involving these verbs. 
Additionally, under the RDV split, we analyze performance on uncommon instrument–verb pairs by grouping all triplet classes according to their $(\text{instrument}, \text{verb})$ pairs and reporting the mean AP within each group.

\paragraph{Embedding Alignment Visualization}
To analyze fine-grained visual–text alignment, we visualize cosine similarity heatmaps between visual tokens and triplet text embeddings. 
Since V-JEPA2 uses a temporal stride of 2, tokens from every two consecutive frames are grouped into one temporal unit. 
After the predictor, cosine similarity between each output token and the triplet text embedding produces one similarity map per unit. 
Each heatmap is overlaid on the second frame of the pair and bilinearly upsampled to the original resolution.


\subsection{Implementation and Training Details}\label{sec:implmentation_details}

We follow the two-stage training strategy of VL-JEPA~\cite{chen2025vl}, on \ours{} In the first stage, the model is trained without trajectory tokens to establish basic visual–language alignment, and in the second stage full model is trained with video clips. ALL video clips used for baseline and \ours{} are constructed by augmenting each frame with six preceding frames. All trainning are done by using 3 RTX Pro 6000 Blackwell GPUs. The text encoder is updated via prompt tuning only with 4 context learnable tokens, and the visual encoder remains frozen. We use AdamW with learning rates of $5\times10^{-5}$ (predictor), $1\times10^{-4}$ (trajectory modules), and $2.5\times10^{-6}$ (text encoder). Finally, we use V-jepa2 \cite{assran2025v} as the visual encoder and initialize the predictor with last 8 layers of llama3.2-1B.
\subsection{Demo Video}

CholecT50 is derived from CholecT80~\cite{twinanda2016endonet} by sampling frames at 1 FPS. 
Accordingly, we train and evaluate under the same 1-FPS RDV split for fair comparison. 
To examine robustness at higher temporal resolution, we additionally provide a demo on an overlapping Cholec80 clip with available ground-truth triplets. 
In this demo, frames are sampled at 10 FPS using a six-frame sliding window, and heatmaps are generated from the top-1 prediction.
\subsection{Results}
\subsubsection{RDV Data Splitting and Unseen Verb Results}
Table~\ref{tab:ivt_combined_results} reports AP and Top@K results under the RDV split and unseen-verb setting to show the overall perfermence of \ours{}. Our \ours{} consistently outperforms all baselines across all metrics. Under RDV, it achieves the best AP$_I$ (86.37), AP$_V$ (59.23), AP$T$ (36.01), and AP${IVT}$ (14.77), as well as the highest $Top@(K{=}|GT|)$ (65.45) and Top@20 (97.02). Table~\ref{tab:verb_given_instrument} further shows that incorporating trajectory information improves recognition of all uncommon triplets, notably “grasper–pack” and “bipolar–grasp”.  In the unseen-verb setting, \ours{} maintains clear gains, reaching AP$_{IVT}$ of 11.26 and the highest $Top@(K{=}|GT|)$ score, demonstrating stronger generalization to unseen verb compositions. Table~\ref{tab:unseen_per_class} also shows consistent per-class AP improvements, particularly for motion-intensive interactions such as tool retraction and fluid aspiration. Table~\ref{tab:verb_given_instrument} shows that \ours{} also achieves gains on less frequent instrument–verb pairs. For example, performance improves from 18.1 to 32.9 for Grasper–pack (+14.8), from 0.2 to 0.8 for Grasper–dissect (+0.6), and from 2.8 to 4.2 for Bipolar–retract (+1.4), demonstrating stronger modeling of uncommon actions.

\begin{table*}[t]
\centering
\footnotesize
\setlength{\tabcolsep}{4pt}
\renewcommand{\arraystretch}{0.95}

\caption{Overall Average Precision (AP) and TOP@K=$|GT|$ under RDV~\cite{nwoye2022data} split and unseen-verb settings. $\ddagger$ represents surgical pretrained VLMs.}
\label{tab:ivt_combined_results}
\vspace{-2mm}
\begin{tabular}{l l cccccccc}
\toprule
\textbf{Setting} & \textbf{Method} & \textbf{AP$_I$} & \textbf{AP$_V$} & \textbf{AP$_T$} & \textbf{AP$_{IVT}$} & \textbf{TOP@($K{=}|GT|$)} & \textbf{TOP@5} & \textbf{TOP@10} & \textbf{TOP@20} \\
\midrule

\multirow{8}{*}{\textbf{Standard RDV}} 
& CLIP-VIT-L-Patch14-FT & 73.63 & 44.43 & 31.67 & 12.62 & 59.32 & 80.32 & 87.04 & 89.60 \\
& CLIP-VIT-L-Patch14-GOAL-FT & 76.04 & 45.93 & 32.08 & 13.15 & 61.07 & 80.93 & 87.99 & 95.04 \\
& SurgVLP-FT$^\ddagger$ & 71.18 & 47.70 & 33.35 & 12.68 & 59.02 & 81.98 & 88.72 & 94.93 \\
& HecVL-FT$^\ddagger$ & 76.07 & 50.92 & 32.09 & 12.67 & 58.03 & 78.80 & 86.59 & 93.58 \\
& PeskaVL-FT$^\ddagger$ & 75.20 & 48.71 & 30.29 & 12.12 & 57.05 & 76.76 & 86.43 & 93.54 \\
& VL-JEPA (Image) & 72.51 & 49.22 & 31.58 & 12.65 & 60.06 & 81.58 & 90.04 & 95.31 \\
& VL-JEPA (Video) & 82.95 & 55.17 & 34.16 & 13.49 & 61.91 & 79.86 & 88.81 & 94.91 \\
& \textbf{TrajPred (Ours)} & \textbf{86.37} & \textbf{59.23} & \textbf{36.01} & \textbf{14.77} & \textbf{65.45} & \textbf{84.33} & \textbf{91.61} & \textbf{97.02} \\

\midrule

\multirow{6}{*}{\textbf{Unseen Verb}} 
& SurgVLP-FT$^\ddagger$ & 54.57 & 35.32 & 21.96 & 8.09 & 15.17 & 22.98 & 29.07 & 33.12 \\
& HecVL-FT$^\ddagger$ & 60.52 & 35.27 & 23.76 & 8.68 & 16.69 & 25.55 & 30.61 & 34.15 \\
& PeskaVL-FT$^\ddagger$ & 57.92 & 35.79 & 23.70 & 8.54 & 17.73 & 26.97 & 31.95 & 34.91 \\
& VL-JEPA (Image) & 53.73 & 32.04 & 23.94 & 8.28 & 17.53 & 25.65 & 32.29 & 34.85 \\
& VL-JEPA (Video) & 62.33 & 35.97 & 24.88 & 9.02 & 19.39 & 27.67 & 32.93 & 35.21 \\
& \textbf{TrajPred (Ours)} & \textbf{74.31} & \textbf{43.62} & \textbf{26.33} & \textbf{11.26} & \textbf{20.88} & \textbf{28.14} & \textbf{32.99} & \textbf{36.27} \\

\bottomrule
\end{tabular}
\vspace{-4mm}
\end{table*}

\begin{table}[t]
\centering
\footnotesize
\setlength{\tabcolsep}{4pt}
\renewcommand{\arraystretch}{0.95}
\caption{Ablation study of \ours{} under the Standard RDV and Unseen-Verb settings. We report TOP@($K{=}|GT|$).}
\begin{tabular}{lcc|c}
\toprule
Setting & Verb Rephrase & Traj Tokens & TOP@($K{=}|GT|$) \\
\midrule
\multirow{4}{*}{\textbf{Standard RDV}}
& $\times$ & $\times$ & 60.51 \\
& \checkmark & $\times$ & 60.23 \\
& $\times$ & \checkmark & 65.37 \\
& \checkmark & \checkmark & \textbf{65.45} \\
\midrule
\multirow{4}{*}{\textbf{Unseen Verb}}
& $\times$ & $\times$ & 17.89 \\
& \checkmark & $\times$ & 19.28 \\
& $\times$ & \checkmark & 19.97 \\
& \checkmark & \checkmark & \textbf{20.88} \\
\bottomrule
\end{tabular}
\label{tab:ablation}
\vspace{-4mm}
\end{table}

\begin{table}[t]
\centering
\vspace{2mm}
\footnotesize
\setlength{\tabcolsep}{4pt}
\caption{Average Precision (AP) on triplets containing uncommon instrument--verb pairs. $\ddagger$ denotes the common verb associated with the corresponding instrument.}
\label{tab:verb_given_instrument}
\vspace{-2mm}
\begin{tabular}{llccc}
\toprule
\multirow{2}{*}{\textbf{Instrument}} & \multirow{2}{*}{\textbf{Verb}} & \textbf{Occurrence} & \textbf{VL-JEPA} & \multirow{2}{*}{\textbf{\ours{}}} \\
& & \textbf{(\%)} & \textbf{(video)}& \\
\midrule
\multirow{4}{*}{Bipolar}
 & coagulate$^{\ddagger}$ & 71.7 & 18.0 & \textbf{18.8} \\
 & dissect   & 20.9 &  0.4 & \textbf{0.9} \\
 & grasp     &  3.1 &  13.7 & \textbf{14.6} \\
 & retract   &  4.3 & 2.8 &  \textbf{4.2} \\
\midrule
\multirow{4}{*}{Grasper}
 & retract$^{\ddagger}$ & 80.8 & 37.8 & \textbf{40.2} \\
 & grasp   & 17.6 & \textbf{14.9} & 13.6 \\
 & dissect &  1.2 &  0.2 & \textbf{0.8} \\
 & pack    &  0.4 & 18.1 & \textbf{32.9} \\
\midrule
\multirow{4}{*}{Hook}
 & aspirate$^{\ddagger}$ & 71.8 & 61.9 & \textbf{63.3} \\
 & irrigate & 12.6 &  4.9 & \textbf{5.4} \\
 & retract  &  8.1 &  4.2 & \textbf{5.2} \\
 & dissect  &  7.6 &  1.1 & \textbf{1.6} \\
\bottomrule
\end{tabular}%
\vspace{-4mm}
\end{table}

\begin{table}[t]
\centering
\caption{Per-class Average Precision (AP) on six selected triplets containing unseen verbs.}
\label{tab:unseen_per_class}
\vspace{-2mm}
\begin{tabular}{lccc}
\toprule
\textbf{Triplet} 
& \textbf{HecVL-FT} 
& \textbf{PeskaVLP-FT} 
& \textbf{\JC{\ours{}}} \\
\midrule
\makecell[c]{``grasper pack\\ gallbladder''} & 0.39 & 0.61 & \textbf{1.02} \\
\makecell[c]{``grasper retract\\ gut''} & 4.79 & 3.82 & \textbf{8.40} \\
\makecell[c]{``grasper retract\\ liver''} & 28.81 & 26.35 & \textbf{38.17} \\
\makecell[c]{``grasper retract\\ omentum''} & 16.84 & 14.26 & \textbf{27.18} \\
\makecell[c]{``irrigator aspirate\\ fluid''} & 12.22 & 14.22 & \textbf{22.49} \\
\makecell[c]{``irrigator irrigate \\ cystic\_pedicle''} & 0.54 & 0.92 & \textbf{3.05} \\
\bottomrule
\end{tabular}
\vspace{-6mm}
\end{table}

\subsubsection{Ablation Study}
Table~\ref{tab:ablation} reports an ablation study of verb rephrasing and trajectory-token conditioning, evaluating the proposed modules built on top of VL-JEPA. Under the Standard RDV setting, trajectory tokens improve TOP@($K{=}|GT|$) from 60.51 to 65.37, while verb rephrasing provides a smaller gain (60.23). The full model achieves the best result of 65.45. Similar improvements are observed in the Unseen-Verb setting, where the full model reaches 20.88.

\subsubsection{Visual features--Text Comparison Results}
\begin{figure*}[t]
\vspace{-2mm}
\centering
\includegraphics[width=2.0\columnwidth]{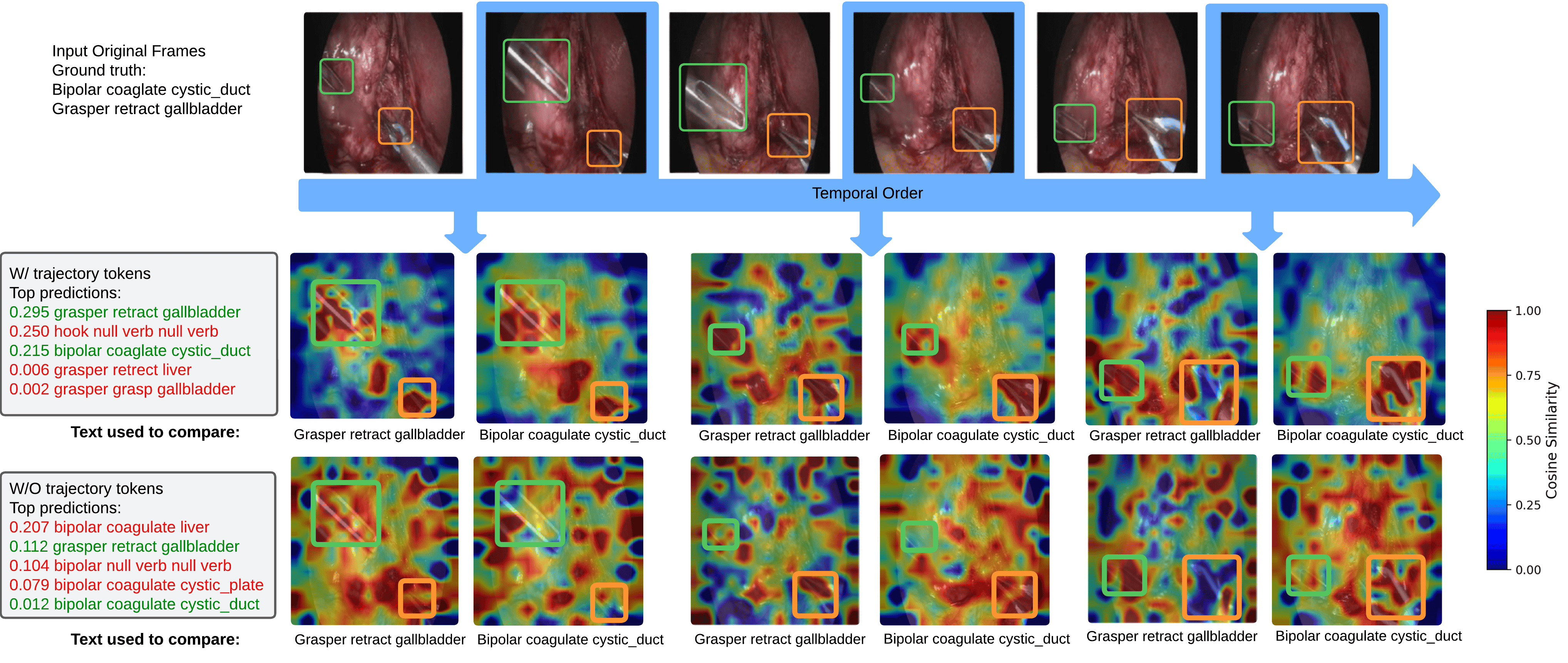}
\vspace{-4mm}
\caption{Image–text similarity heatmaps over six consecutive frames, where red indicates higher cosine similarity. Similarity is computed between predictor-generated spatiotemporal tubelet tokens and the embedding of the ground-truth triplet (when it appears in the top-5 predictions). Each heatmap represents a two-frame temporal unit and is overlaid on the second frame. Top-5 triplet scores are shown alongside the heatmaps. Green and orange bounding boxes denote the grasper and bipolar instruments, respectively.}
\label{fig_heatmap}
\vspace{-3mm}
\end{figure*}

We report visualization heatmaps of cosine similarity between visual features and text embeddings to show that \ours{} produces representations that better localize action-relevant regions.
Fig.~\ref{fig_heatmap} compares the prediction performance and heatmap quality between \ours{} and the baseline VL-JEPA (without trajectory tokens). Regions with high similarity indicate where the predictor’s visual tokens are most semantically aligned with the surgical action. With trajectory tokens, the highlighted regions concentrate on the active instrument and its interaction site, suggesting that the predictor learns more precise action representations. Moreover, the ground-truth triplets achieve higher confidence scores within the top-5 predictions (e.g., “grasper retract gallbladder” 0.295 and “bipolar coagulate cystic-duct” 0.215). In contrast, without trajectory tokens, the heatmaps become diffuse and often shift toward endoscope boundaries, likely reflecting camera motion or background patterns rather than meaningful interactions. Correspondingly, the ground-truth confidence scores drop significantly (0.112 and 0.012). Notably, both triplet heatmaps highlight the two instruments in the scene, which we attribute to the unmasked self-attention in the predictor that enables bidirectional interactions among triplet tokens.

\begin{figure}[t]
\centering
\includegraphics[width=0.9\columnwidth]{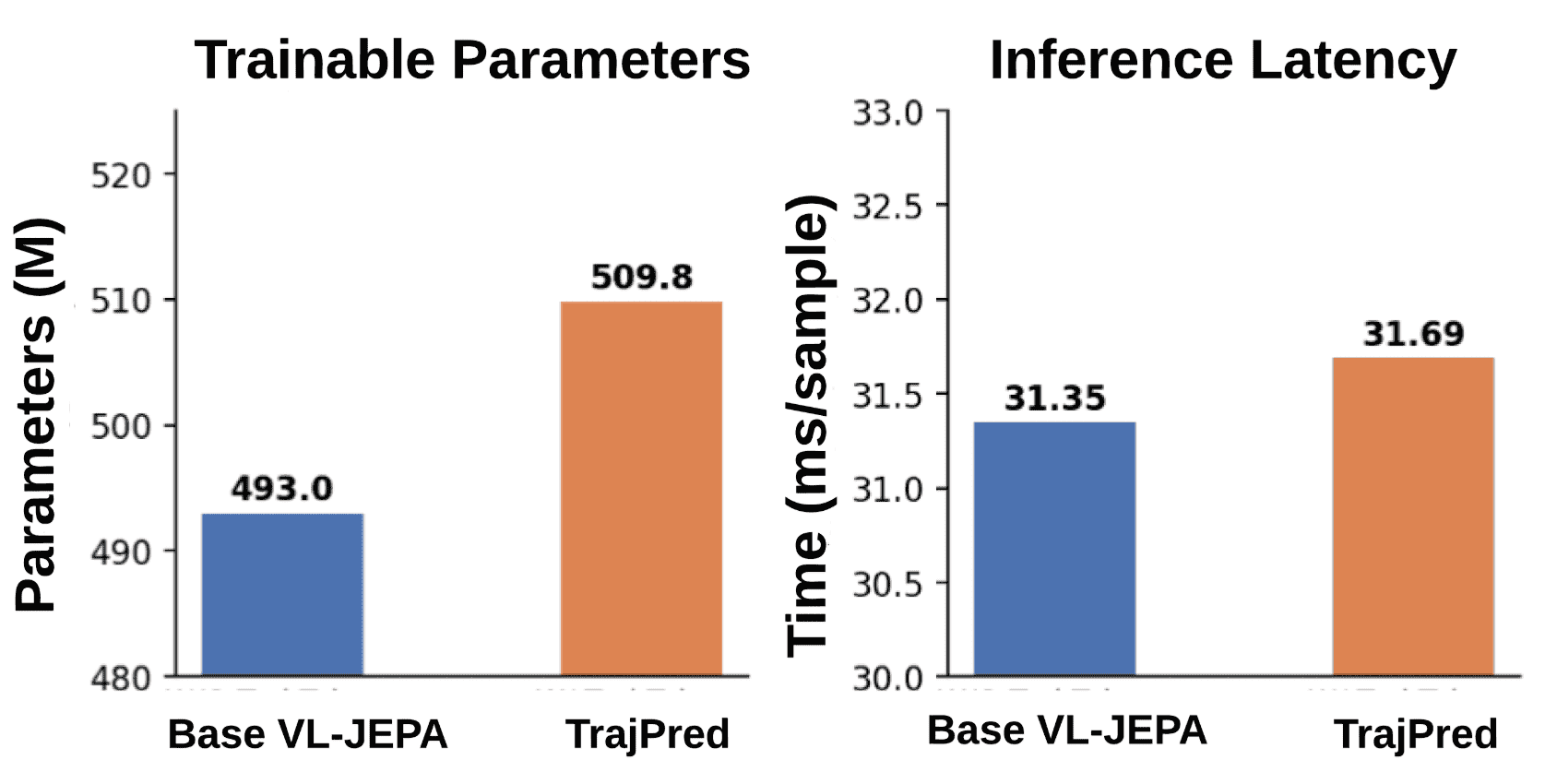}
\vspace{-4mm}
\caption{\JC{Comparison of parameter count and inference latency }}
\label{fig_overhead}
\vspace{-6mm}
\end{figure}

\section{Discussion and Future Work}
The main contribution for verb rephrasing technique mainly lies on the unseen verb setting. Verb rephrasing technique improves overall performance in the Unseen-Verb setting, as shown in Table~\ref{tab:ablation}. The triplet “grasper pack gallbladder” exhibits very low performance across all models because it is an extremely long-tailed action (Table~\ref{tab:unseen_per_class}). Even in this case, verb rephrasing still provides a small performance improvement. This suggests that alternative linguistic expressions help the model associate unseen contexts with semantically related textual descriptions learned during training. Such an approach can be particularly useful in the surgical domain, where many terms are specialized and less common compared to general-domain vocabulary.

In Table~\ref{tab:verb_given_instrument}, we report the AP scores for triplets where each instrument performs both its common and uncommon actions to examine whether trajectory information helps distinguish different instrument behaviors, as instruments may perform multiple actions. Notably, incorporating trajectory information enables \ours{} to achieve improvements on these rare triplets. This suggests that explicitly encoding instrument trajectories helps capture finer details of instrument behavior. 

In terms of efficiency, the trajectory token module in \ours{} introduces minimal computational overhead. (Fig.~\ref{fig_overhead}). It increases model size by only 3.4\ (493.0M → 509.8M) and adds at most four trajectory tokens during inference, resulting in negligible extra self-attention cost. End-to-end latency rises by just 0.34 ms per sample (31.35 → 31.69 ms, +1.1\%). This demonstrates that our trajectory tokens encoding path provides improvement without adding much computational burden.

Finally, we use 1 FPS sampling to capture meaningful motion while reducing redundancy. However, surgical actions vary widely in speed—from rapid cutting to subtle tissue retraction—fixed frame sampling may fail to capture reliable motion cues for all action type; future work will explore adaptive frame rates and extend this spatiotemporal encoding strategy to large-scale pretraining for improved robustness and generalization.

\section{Conclusion}
In this work, we introduce \ours{}, a predictive embedding framework for surgical VLMs conditioned on instrument trajectory information for surgical action recognition. 
To address the limited use of temporal information across frames, \ours{} leverages video clips and incorporates trajectory tokens to explicitly encode instrument motion. 
This trajectory-conditioned prediction further mitigates the loss of fine-grained details often caused by contrastive learning alignment. 
Experimental results and qualitative visualizations demonstrate improved performance and stronger alignment between image patches and textual representations. 
These findings highlight the importance of modeling detailed motion cues beyond high-level image- or clip-level vision–language alignment for surgical understanding.






\end{document}